# AN EFFICIENT METRIC OF AUTOMATIC WEIGHT GENERATION FOR PROPERTIES IN INSTANCE MATCHING TECHNIQUE

Md. Hanif Seddiqui[1], Rudra Pratap Deb Nath[1] and Masaki Aono[2]

[1]Department of Computer Science and Engineering, University of Chittagong, Chittagong-4331, Bangladesh

[2]Department of Computer Science and Engineering, Toyohashi University of Technology, Toyohashi, Aichi 441-8580, Japan

## ABSTRACT

*The proliferation of heterogeneous data sources of semantic knowledge base intensifies the need of an automatic instance matching technique. However, the efficiency of instance matching is often influenced by the weight of a property associated to instances. Automatic weight generation is a non-trivial, however an important task in instance matching technique. Therefore, identifying an appropriate metric for generating weight for a property automatically is nevertheless a formidable task. In this paper, we investigate an approach of generating weights automatically by considering hypotheses: (1) the weight of a property is directly proportional to the ratio of the number of its distinct values to the number of instances contain the property, and (2) the weight is also proportional to the ratio of the number of distinct values of a property to the number of instances in a training dataset. The basic intuition behind the use of our approach is the classical theory of information content that infrequent words are more informative than frequent ones. Our mathematical model derives a metric for generating property weights automatically, which is applied in instance matching system to produce re-conciliated instances efficiently. Our experiments and evaluations show the effectiveness of our proposed metric of automatic weight generation for properties in an instance matching technique.*

## 1. INTRODUCTION

With the rapid growth of diversified heterogeneous semantically linked data, often called as instances, instance matching becomes a key factor to reconcile the data. In semantic web, instances of people, places and things, are connected by means of concepts, properties and their instantiation in domain ontologies. However, ontologies in a same domain are often defined differently by different creators influenced by their interest, social behaviours and after all due to their different needs. That imposes a challenge to reconcile instances to integrate information of semantic knowledge bases.

A semantic knowledge base contains assertion about instances of two disjoint sets called "concepts", $C$ and "relations", $R$ which is technically called as *property* in Resource Description Framework (RDF) [1] and in Web Ontology Language (OWL) [2]. The semantic knowledge base is defined in [3] as follows:

$$KB=(C,R,I, _{C},\ _{R})$$ (1)

With the definition of the knowledge base, we find that it consists of two disjoint sets $C$ and $R$ as defined before, a set $I$ whose elements are called instance identifiers, a function, $_{C}$: $C$ $\mathfrak{R}(I)$ called concept instantiation and a function $_{R}$: $R$ $\mathfrak{R}(I^2)$ with $_{R}(r) \subseteq _{C}(dom(r))$ x $_{C}(ran(r))$, for all $r \in R$. The function $_{R}$ is called relation instantiation.

Currently a large number of ontology instances are available in semantic knowledge bases: *AllegroGraph*[1] [4] contains more than one trillion triples, *a basic building block of Semantic Web formed as <subject> <predicate> and <object>*, Linked Open Data (LOD) [5] contains more than fifty billion triples and there are more other knowledge bases too like *DBpedia* [6], DBLP [7] and so on. Moreover, several individual groups are also working to create billions of triples to represent ontology instances of semantic web. Due to the proliferation of semantically connected instances, automatic instance reconciliation is getting researchers' attention. The problem of instance reconciliation is often called as instance matching problem.

Ontology instance matching is a relatively new domain for researchers in comparison to record linkage, which has a classical state-of-the-art [8, 9, 10], although there is a close relationship between instance matching to record linkage. Instance matching is an important approach to connect all the islands of instances of semantic web to achieve the interoperability and information integration issues. Instances in knowledge base contain descriptions through a number of properties.

The description of instances varies in their natural language based lexicon, in their structure and so on. For example, a person's name is differently described across nations and even in the citation of different publications, and date has also wide variants. Therefore, instance matching becomes a formidable task for measuring the proximity considering different transformations in their descriptions. There are three basic transformations across instances: value transformation, logical transformation and structural transformation. Value transformation focuses on the description variation in their lexicon, while logical transformation is about the typeset variations in terms of ontology concept. However, the structural variation is more challenging as instance functionality varies in terms of ontology properties. To cope with the missing information of instances in different knowledge base is also challenging. In addition, instances from different ontology impose some extra challenges as we need ontology schema alignment before going for instance matching.

As in equation 1 of the definition of knowledge base, instances are well defined in terms of properties. Properties are classified into *DatatypeProperty*, where the range of the property is a literal and *ObjectProperty*, where the range of the property is another instance. An instance may be defined by instantiating properties from a few numbers to hundreds of them. Every property may have different impact on their associated instances. This imposes additional challenges in instance matching techniques.

Most of the instance matching research has been focused on the straight forward instance matching problem with different type of transformations. In the contrary, HMatch(I) [11] tried to address automatically detecting property weight. However, it was only focusing on the distinct value based weight generation, which has a negative impact when there are a small number of available instances containing that property. This paper, in fact, addresses the necessity of weight

---

of properties, where the authors refer to as featuring properties. Let us give a comprehensive short example. An instance of type "Person" may have property values attached to *hasEmail* and *hasAge* properties. Once, two instances have same values for *hasEmail* property. Both of the instances should be same even if the values against *hasAge* are different as the data might be captured at different year. Therefore, *hasEmail* and *hasAge* have different weight factor. Apparently, *hasGender* may have different weight factor than *hasEmail* and *hasAge* to identify an instance.

In [12] authors propose that properties, which have a maximum or an exact cardinality of 1 have a higher impact factor on the matching process. However, it has a fallacy in logic. For instance, a person has exactly one father i.e. the cardinality of *hasFather* property is one. However, father of all siblings is the same one. Therefore, the system may falsify that two persons are same if they have same value for the property *hasFather*. So far researches in Ontology Instance Matching (OIM) assigns the weight factor to the property in top down approach i.e. by either analysing the schema of the ontology or manually. However, our effort is to automatically impose the weight by analysing the information of instances which is more convincing and practical.

We investigate different factors that affect weight of a property. Eventually we find three factors: the uniqueness of the property values, the number of instances a property contains, and the total number of instances in the knowledge base. Obviously, the uniqueness of property values has the direct relationship with property weight. Combining the three factors we find that property weight is directly proportional to the ratio of the number of distinct values of a property to the number of instances contain that property. This, in turn, depicts that the number of instances contained a property has a negative effect if the number of distinct values is constant. Moreover, property weight is also non-linear proportional to he ratio of the number of distinct values of a property to the number of total instances. This, in turn, gives us message that the total number of instances has a negative effect if the number of distinct values is kept constant; however the total number of instances is increased. Therefore, measuring a straight forward property weight by linear equation may not work properly. Suppose, out of one million instances only ten of them contain birth-date, and unfortunately all of them are unique. In that case, it is not wiser to consider that one million instances must contain unique birth-date. Therefore, we propose a metric combining the factors together to generate relatively effective weight factors.

In this regard, we experimented with the proposed metric of property weight generation applied in our previous core instance matching technique [13]. The result depicts that our proposed metric for property weight generation has better impact over instance matching technique.

Ontology instance matching is required to compare different individuals with the goal of recognizing the same real-world objects. In particular, the application of instance matching plays an important role in information integration, identity recognition and in ontology population. Ontology schema matching and instance matching work in each other to facilitate to discovering semantic mappings between possibly distributed and heterogeneous semantic data. Identity recognition is a widely used term in database and emerging topic in the semantic web of detecting whether two different resource descriptions refer to the same real-world entity, namely an individual. Ontology population is evolved by acquiring new semantic descriptions of data extracted from heterogeneous data sources. For this ontology population, instance matching plays a crucial role to correctly perform the insertion activity and to discover a set of semantic mappings between a new incoming instance and the set of instances already stored in an ontology.

The rest of the paper is organized as follows. **Section 2** compares our idea with other existing related work to articulate a research gap. The factors that affect the property weight are articulated at **Section 3** along with some comprehensive examples. **Section 4** contains the mathematical explanation of our metric and necessity of the different considerable factors. The detail implementation of our instance matching technique along with our integrated metric to generate property weight factors are described at **Section 5**. **Section 6** includes experiments and evaluation to show the effectiveness of our proposed metric to generate property weight factors to match different instances. Concluded remarks and some future directions of our work are described in **Section 7**.

## 2. RELATED WORK

The rising demand of sharing knowledge, data and information within same or heterogeneous knowledge bases has recently attained a novel attention on issues related to ontology and instance matching. Until now, many researchers have invested their efforts on ontology instance matching to resolve the interoperability issues across heterogeneous sources. In SERIMI [14], instances are matched between a source and target datasets, without prior knowledge of the data, domain or schema of these datasets. However, in the instance matching process, SERIMI does not impose any weights to the properties associated with instances. The weight of each property can be manually specified by a domain expert [15] and [11] or it can be automatically determined through statistical analysis [16], [17], [18]. In HMatch 2.0 [11], each property is associated with a weight ranging from 0 to 1 expressing the capability of the property for the goal of equivocally identifying the individual in the domain of interest. This weight is defined during the featuring properties identification step of the instance matching process. In BOEMIE, property weights are manually defined for the considered domain by taking into account the results of the extraction process from a corpus of (manually) annotated multimedia resources. Nonetheless, manual definition of weight requires involvement of domain experts and the definition of weight may vary among different domains.

To discover semantic equivalence between persons in online profiles or otherwise, an appropriate metric is proposed in [12] for weighting the attributes which are syntactically and/or semantically matched. The properties that have a maximum or an exact cardinality of 1 have a higher impact on the likelihood those two particular profiles are semantically equivalent. However, it has a fallacy in logic. For instance, a person has exactly one father i.e. the cardinality of *hasFather* property is one. However, father of all siblings is the same one. Therefore, the system may falsify that two persons are same if they have same value for the property *hasFather*.

A further refinement of the instance matching process is taken into account considering the frequency of each value occurs [16] in the knowledge base. In particular, a pair of matching attribute values will receive a high weight if these values occur with a low frequency within the domain, while they will receive a low weight otherwise.

RiMOM [19] used several instance matching benchmark data sets to evaluate their systems namely, A-R-S, T-S-D and IIMB. However, for different datasets, their matching strategy is different.

In [20] and [21], J. Huber et. al. have proposed CODI: Combinatorial Optimization for Data Integration in where they emphasize on object-properties to determine the instances for which the similarity should be computed. Although object-properties have a strong influence in the matching process, involvement of data-properties in the matching process is also necessary.

Till date researchers in the domain of ontology instance matching tried to assign the weight factor to the property in a top down manual approach. In this approach, researchers were assigning weight factors to the property either by analysing the schema of the ontology manually or by domain experts arbitrarily. However, our effort is to automatically generate the weight by analysing the information of instances which is more convincing, generic and practical.

## 3. FACTORS THAT INFLUENCE THE PROPERTY WEIGHT

We have factorized the property weight considering very classical information theoretic approaches. The basic intuition behind the use of the this approach is that the more probable a concept is of appearing then the less information it conveys, in other words, infrequent words are more informative than frequent ones.

Information theoretic approaches are well defined in a couple of research works by [22, 23, 24, 25, 26]. They obtain their needed Information Content (IC) values by statistically analysing corpora. They associate probabilities to each concept in the taxonomy based on word occurrences in a given corpus. The IC value is then obtained by considering the negative log likelihood [24, 27]:

$$ic_{res} (c) = -\log p(c), \qquad (2)$$

where *c* is any concept and *p(c)* is the probability of encountering *c* in a given corpus. [24] was the first to consider the use of this formula, that stems from the work of Shannon [28], for the purpose of semantic similarity judgments.

Moreover, instances in knowledge base contain values associating with properties. Some properties like *name*, *date-of-birth*, *and homepage* have larger weighting factors than the properties like *height*, *frequency* and so on. However, determining the weight factor automatically is a formidable task. Some properties have great influence on identifying instances, while the other has less influence in a semantic knowledge base. For example, an instance of type *Person* may have property values attached to *hasEmail* and *hasAge* properties. Once, two instances have same values for *hasEmail* property. Both of the instances must be same even if the values against *hasAge* are different as the data is captured at different year. Therefore, *hasEmail* and *hasAge* have different weight factor.

The above fact depicts from Equation 2 and we consider that properties with distinct values are having more weight than that of a property with duplicate values. Therefore, duplicate values are influencing weights as a negative factor.

### 3.1. Influence of Negative Factors

Our basic hypothesis to identify the influence of a property on instance identification is that a property has higher weight if its values do not repeat in a semantic knowledge base like a primary key in a database repository. Alternatively a property has less weight if it repeats in the knowledge base. As many times the property value repeats, it loses its ability to identify an instance.

If a property value repeats, the weight is penalized by a negative probability factor, *np* defined as a ration of the number of repetition to the number of instances the property belongs to, i.e.

$$np_1 = \frac{Number\ of\ value\ repetition\ of\ property\ p}{Number\ of\ instances\ contain\ property\ p} \tag{3}$$

$$= \frac{|dup|}{|i \ni p|}$$

Moreover, the ratio of the property value repetition to the total number of instance has also negative effect on property weight. Primarily, let us consider the fact in the equation below:

$$np_2 = \frac{Number\ of\ value\ repetition\ of\ property\ p}{Total\ number\ of\ instances} \tag{4}$$

$$= \frac{|dup|}{|I|}$$

## 3.2. Our Proposed Property Weight Factors

As described in Subsection 3.1, there are two types of negative factors associated with our automatic weight generation for properties of semantic knowledge base, namely $np_1$ and $np_2$ and they are defined as primarily as follows:

$$np_1(p) = \frac{|dup|}{|i \ni p|}, \tag{5}$$

$$np_2(p) = \frac{|dup|}{|I|}, \tag{6}$$

where $|dup|$ is the number of value duplication for a property $p$ available throughout the knowledge base and $i$ represents an instance. Moreover, $|i \ni p|$ represents the total number of instances containing the property, $p$ and $|I|$ represents the total number of instances in the knowledge base.

The probability of identifying an instance with $p$ would be denoted as $Prob(p)=1-np(p)$. We consider the probability as the weight of that property. Therefore, we measure the weight of each property of an ontology schema used in a knowledge base as a joint probability and is stated as follows:

$$\text{weight (p)} = (1.0 - np_1\ (p)) \quad (1.0 - np_2\ (p)), \tag{7}$$

where $np_1$ and $np_2$ are defined above.

## 4. MATHEMATICAL EXPLANATION

The primary equation 7 articulates the impact of negative factors in terms of the number of value repetition of a property. However, our metric concentrates on the reverse of the value repetition, i.e. the number of distinct value of a property available in a knowledge base. The following subsection focuses on the mathematical derivations and reasoning.

## 4.1. Mathematical Derivations

Let us start from the joint probability equation 7 for mathematical derivation and to look insight the nature of the equation.

$$weight(p) = (1.0 - np_1(p)) * (1.0 - np_2(p))$$

$$= \left(1.0 - \frac{|dup|}{|i \ni p|}\right) * \left(1.0 - \frac{|dup|}{|I|}\right)$$

$$= \frac{|i \ni p| - |dup|}{|i \ni p|} * \frac{|I| - |dup|}{|I|}$$

$$= \frac{|i \ni p| - |dup|}{|i \ni p|} *$$

$$\frac{(|I| - |i \ni p|) + (|i \ni p| - |dup|)}{|I|}$$

$$= \frac{|distinct|}{|i \ni p|} * \frac{(|i \not\ni p|) + |distinct|}{|I|}$$

$$[\because |i \ni p| - |dup| = |distinct|, and$$
$$(|I| - |i \ni p|) = (|i \not\ni p|)]$$

(8)

where *|distinct|* is the number of instances that contain distinct values to property *p*. Although the first term *|distinct|²/(|I ⋑ p|\*/I|)* is quite convincing in equation 8, however in the second term *(|distinct|\*/I ⋫ p|)/(|I ⋑ p|\*/I|)*, we consider that *|I ⋫ p|* has a positive contribution to the weight factor, which is a contradiction. Let us consider that there are 1 million instances as human being in a knowledge base and 10 instances of them are containing *date-of-birth* property values and unfortunately all of them have a distinct value. This obviously does not guaranty that the rest instances will contain distinct values. On the other hand, it is not also guaranteed that most of them are duplicate value. Therefore, we need a factorization parameter, $\lambda$ before the second term as a multiplier, which is defined as below:

$$\lambda = \begin{cases} \frac{|distinct|}{|i \ni p|} & : \frac{|i \ni p|}{|I|} \geq \delta \\ sigmod(\frac{|i \ni p|}{|I|}) * \frac{|distinct|}{|i \ni p|} & : otherwise \end{cases}$$

(9)

where $\delta$ is the empirical threshold and *sigmod* is a logistic distribution function defined below:

$$sigmod(x) = \frac{1}{1 + e^{-(x-\mu)/s}}$$

(10)

where *s* and $\mu$ are two empirical constants as defined to control the distribution as starting closely from 0.5 and ending around 0.95 for the argument parameter *(|i ⋑ p|)/|I|* in our experiment. Although *s* is a scaling parameter, we define the value of *s* as 0.2 to set the maximum value of the *sigmod* function at around 0.95. On the other hand, although $\mu$ is a location parameter to set the center of origin, we set the value at 0.1 to achieve the minimum value of the *sigmod* function at 0.5.

$$weight(p) = \frac{|distinct|}{|i \ni p|} * \frac{\lambda * (|i \not\ni p|) + |distinct|}{|I|}$$

(11)

Thereafter the equation 8 becomes as:

## 4.2. Comprehensive Example

Let us consider a number of comprehensive examples to understand the equation 11.

In equation 11, if we consider that a property, *p* is densely instantiated among instances, i.e. every instance in the knowledge base contains some values of *p*, then *|i ⋫ p|* is zero. Hence, the equation 11 becomes:

$$weight(p) = \left(\frac{|distinct|}{|i \ni p|}\right) * \left(\frac{|distinct|}{|I|}\right)$$

(12)

Let the total number of instance, $|I|$ and the number of instances having distinct values $|distinct|$ of a property, $p$ be constants. In this case, if the number of instances containing $p$, denoted as $|i \ni p|$ increases, the possibility of identifying an instance with that property decreases. Because, as the $|distinct|$ remain constants and $|i \ni p|$ increases, therefore, duplicate values increases, which means probability of identifying an instance decreases and hence weight of the property decreases and vice-versa. This scenario depicts the natural effect and is successfully addressed in equation 11.

Let the total number of instance, $|I|$ and the number of instances containing property $p$ denoted by $|i \ni p|$ be constants. In this case, if the number of distinct values denoted as $|distinct|$ increases, the possibility of identifying an instance with that property increases. Therefore, duplicate values decreases, which means probability of identifying an instance increases and hence weight of the property increases and vice-versa. This scenario also depicts the natural effect and is successfully addressed in equation 11.

Let the number of distinct values, denoted as $|distinct|$ and the number of instances containing property $p$ denoted by $|i \ni p|$ be constants. In this case, if the total number of instance, denoted as $|I|$ increases, the possibility of identifying an instance with that property decreases as the non-identifiable instances increase, which means probability of identifying an instance decreases and hence weight of the property decreases and vice-versa. This is also addressed in equation 11.

In equation 12, if we consider that a property, $p$ is sparsely instantiated among instances, i.e. some instances in the knowledge base may not contain values of $p$, then $|i \not\ni p|$ is not zero. Hence, the equation 11 remains as it is.

Let the total number of instance, $|I|$ and the number of instances having distinct values $|distinct|$ of a property, $p$ be constants. In this case, if the number of instances containing p, denoted as $|i \ni p|$ increases, the possibility of identifying an instance with that property decreases. Because, as the $|distinct|$ remain constants and $|i \ni p|$ increases, therefore, duplicate values increases, which means probability of identifying an instance decreases and hence weight of the property decreases and vice-versa.

Let the total number of instance, $|I|$ and the number of instances containing property $p$ denoted by $|i \ni p|$ be constants. In this case, if the number of distinct values denoted as $|distinct|$ increases, the possibility of identifying an instance with that property increases. Therefore, duplicate values decreases, which means probability of identifying an instance increases and hence weight of the property increases and vice-versa.

Let the number of distinct values, denoted as $|distinct|$ and the number of instances containing property $p$ denoted by $|i \ni p|$ is constants. In this case, if the total number of instance, denoted as $|I|$ increases, $\lambda * |i \not\ni p|$ increases partly, therefore the possibility of identifying an instance with that property decreases as the non-identifiable instances increase, which means probability of identifying an instance decreases and hence weight of the property decreases and vice-versa. Therefore, it is now obvious that the term $\lambda * |i \not\ni p|$ does not affect the natural behaviour, rather it only reduces the adverse effect of $|i \not\ni p|$.

### 4.3. Quantity Normalization

As the number of instances $|I|$, the number of distinct values of a property $p$ denoted by $|distinct|$, the number of instances contain property $p$ denoted by $|i \ni p|$ and the number of instances that does not contain $p$ and is denoted by $|i \not\ni p|$ are usually degree of large numbers, therefore we consider using log of the terms in equation to reduce the adverse effect of numbers. Hence, the equation 11 becomes:

$$weight(p) = \left( \frac{\log(|distinct| + 1)}{\log(|i \ni p| + 1)} \right) * \qquad (13)$$
$$\left( \frac{\log(|distinct| + 1) + \lambda * \log(|i \not\ni p| + 1)}{\log(|I| + 1)} \right)$$

## 5. OUR INSTANCE MATCHING SYSTEM

Our primitive instance matching system [13, 29] did not get the essence of automatic weight generation. However, we still consider the system as a core of our augmented approach.
Our primitive system of instance matching contains: 1. Ontology Alignment Module, 2. Semantic Link Cloud (SLC) Generation module, and 3. Instance Matching Algorithm.

### 5.1. Ontology Alignment

A concept is neither complete nor explicit in its own words. Therefore, concepts are organized in a semantic network or taxonomy associated with a number of relations to define them explicitly for avoiding polysemy problem. Our ontology schema matching algorithm [30, 13, 31] takes the essence of the locality of reference by considering the neighbouring concepts and relations to align the entities of ontologies.

Our algorithm of ontology alignment starts off a seed point called an anchor, where the notion anchor is a pair of "look-alike" concepts from each of two ontologies. Starting off an anchor point our scalable algorithm collects two sets of neighbouring concepts across ontologies. As our algorithm starts off an anchor and explores to the neighbouring concepts, it does not depend much on the sizes of the ontologies. Thus, our algorithm has a salient feature of size independence in aligning ontologies. Our algorithm achieves enhancement in terms of scalability and performance in aligning large ontologies.

### 5.2. SLC Generation Module

Semantic Link Cloud (SLC), collection of linked information of an instance is an important step toward the instance matching. Users often describe an instance in different ways and even by different, however, neighbouring concepts of an ontology. This often leads to undetected or misaligned pairs. Collection of semantically linked resources of ABox along with concepts or properties of TBox specifies an instance at sufficient depth to identify instances even at a different location or with quite different label. Therefore, our proposed method collects all the linked information from a particular instance as a reference point. The linked information is defined as the concepts, properties or their values which have a direct relation to the reference instance, and is referred to a semantic link cloud.

## 5.3. Instance Matching Algorithm

The strength of our instance matching algorithm depends mainly on the efficiency of generation of SLC, and ontology schema matching.

```
Algo.  instanceMatch (ABox ab₁,
                      ABox ab₂, Alignment A)
1. for each insᵢ ∈ ab₁
2.    slcᵢ=generateSLC(insᵢ, ab₁)
3.    for each insⱼ ∈ ab₂
4.       slcⱼ=generateSLC(insⱼ, ab₂)
5.       if ∃a(c₁,c₂) ∈ A|c₁ ∈ Block(ins₁.type) ∧
                        c₂ ∈ Block(ins₂.type)
6.          if IA(slcᵢ, slcⱼ) ≥ δ₂
7.             imatch=imatch ∪ makeAlign(insᵢ, insⱼ)
```

*Figure 1: Pseudo code of the Instance Matching algorithm.*

The algorithm in Fig.1 portrays a simple flow of the matching algorithm. For an SLC of an instance is matched against every SLCs of instances of knowledge base (line 1 through 4 in Fig.1) if and only if there is an aligned concepts across *Block (ins1.type)* and *Block (ins2.type)* (as there exists a condition at line 5 in Fig.1). *Block (concept)* is a related concept block and *generateSLC(ins, ab)* collects an SLC against an instance *ins* in ABox *ab*. An SLC usually contains concepts, properties, and their consolidated values. Every value of an SLC is compared with that of another SLC (as of line 6 of Fig.1) by affinity measurement metric to calculate similarity between two SLCs. Once similarity value is greater than the threshold, it is collected as an aligned pair (as stated at line 7 in Fig.1). Finally, the algorithm produces a list of matched instance pairs.

Given two individuals $i_1$ and $i_2$ that are instances of the same (or aligned) concept, the instance affinity function $IA(i_1, i_2)$ [0, 1] provides a measure of their affinity in the range *[0,1]*. For each pair of instances, instance affinity, *IA* is calculated by taking all the properties, their values and other instances of the pair of SLCs into account.

## 5.4. Automatic Weight in Instance Matching System

We augmented our system by introducing a primitive automatic weight generation technique [17]. We further improve our primitive automatic weight generation technique with our proposed metric of automatic weight generation [32]. The overall augmented instance matching system is depicted in Fig. 2.

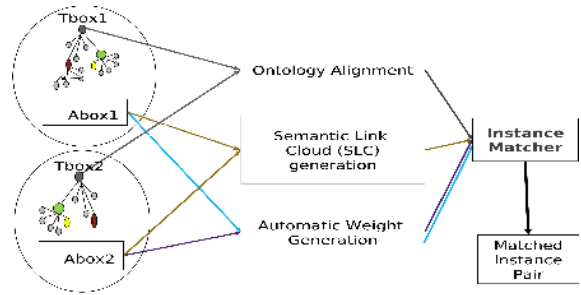

*Figure 2: Overall system with our proposed metric of automatic weight generation.*

Considering weight factor assigned to each of the property automatically, we define the affinity between two SLCs by modified affinity measurement metric as follows:

$$IA_{pf}(slc_1, slc_2) = \frac{\sum_{p \in slc_1, q \in slc_2} E(p, q) \cdot (W_p + W_q)}{\sum_{p \in slc_1} W_p + \sum_{q \in slc_2} W_q - \gamma} \qquad (14)$$

where $\gamma$ represents the factors for missing property values.

## 6. PERFORMANCE EVALUATION

We perform a number of experiments on IIMB data sets of 2009 and 2010 versions and evaluated with evaluation metrics.

### 6.1. Data sets

A generated benchmark to test the efficiency of an instance matcher is called as ISLab Instance Matching Benchmark (IIMB)[2]. The test-bed provides OWL/RDF data about actors, sport persons, and business firms.

We have used two different versions of IIMB datasets: 2009 version and 2010 version. In 2009 version, the main directory contains 37 sub-directories and the original ABox and the associated TBox (abox.owl and tbox.owl). The original ABox contains about 222 different instances with a number of associated property values. Each sub-directory contains a modified ABox (abox.owl + tbox.owl) and the corresponding mapping with the instances in the original ABox (refalign.rdf). The benchmark data is divided into four major groups: value transformation (001-010), structural transformation (011-019), logical transformation (020-029) and combination transformation (030-037) [33].

The 2010 edition of IIMB is a collection of OWL ontologies consisting of 29 concepts, 20 object properties, 12 data properties and thousands of individuals divided into 80 test cases. In fact, in IIMB 2010,80 test cases are defined and divided into 4 sets of 20 test cases each. The rest three sets are different implementations of data value, data structure and data semantic transformations, respectively, while the fourth set is obtained by combining together the three kinds of transformations. IIMB 2010 is created by extracting data from Freebase, an open knowledge base that contains information about 11 million real objects including movies, books, TV shows, celebrities, locations, companies and more. The benchmark has been generated in a small version consisting in 363 individuals and in a large version containing 1416 individuals [34]. Here, large version set is considered in evaluation.

We perform two independent experiments for our instance matcher by not considering property weight and considering property weight on the IIMB benchmark data set. The consecutive sections contain the corresponding evaluation respectively.

---

2     http://islab.dico.unimi.it/iimb/

## 6.2. Evaluation Metrics

In the experiment of instance matching, we have conducted evaluations in terms of *precision*, *recall* and *f-measure* as defined below:

- *Precision, P*: It is the ratio of the number of correct discovered aligned pairs to the total number of discovered aligned pairs.
- *Recall, R*: It is defined as the ratio of the number of correct discovered aligned pairs to the total number of correct aligned pairs.
- *F-Measure*: It is a measure to combine precision, *P* and recall, *R* as *(2 * P * R)/(P+R)*.

## 6.3. Without Weight Factors

For the first time, an instance matching track was proposed to the participants in the Ontology Alignment Evaluation Initiatives, 2009[3]. Our primitive instance matching algorithm produces results on IIMB datasets of 2009 without considering property weight in OAEI campaign [13]. The result is portrayed at Table 1.

Table 1. Instance matching results against IIMB benchmarks at OAEI-2009 without weight factor

| Datasets | Transformation | Prec. | Rec. | F-Measure |
|----------|----------------|-------|------|-----------|
| 001-010 | Value transformations | 0.99 | 0.99 | 0.991 |
| 011-019 | Structural transformations | 0.72 | 0.79 | 0.751 |
| 020-029 | Logical transformations | 1.00 | 0.96 | 0.981 |
| 030-037 | Several combinations of the previous transformations | 0.75 | 0.82 | 0.786 |

Moreover, Fig. 3 demonstrates the results of the participants [33] of OAEI-2009 in where *AFlood* is our instance matcher without considering property weight.

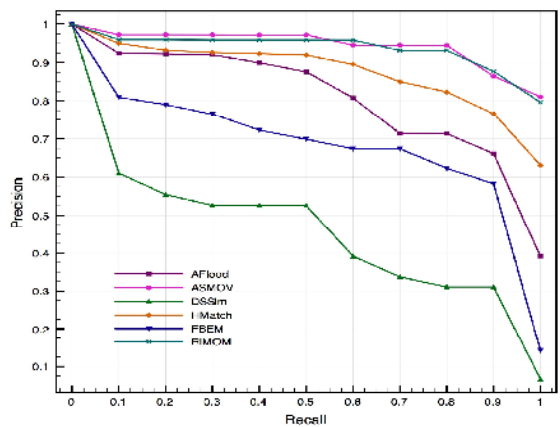

*Figure 3: Instance matching results against IIMB benchmarks. Instance matcher called AFlood is our previous algorithm without considering weight factors.*

---

3       http://oaei.ontologymatching.org/2009/

## 6.4. With Primitive Weight Factors

Our system with primitive weight generation technique of instance matching shows its strength over our basic instance matching system without weight factor [17]. The result is depicted in Table 2.

Table 2. Instance matching results against IIMB benchmarks considering primitive weight factors

| Datasets | Transformation | Prec. | Rec. | F-Measure |
|---|---|---|---|---|
| 001-010 | Value transformations | 1.00 | 1.00 | 1.000 |
| 011-019 | Structural transformations | 0.89 | 0.81 | 0.848 |
| 020-029 | Logical transformations | 1.00 | 1.00 | 1.000 |
| 030-037 | Several combinations of the previous transformations | 0.96 | 0.82 | 0.840 |

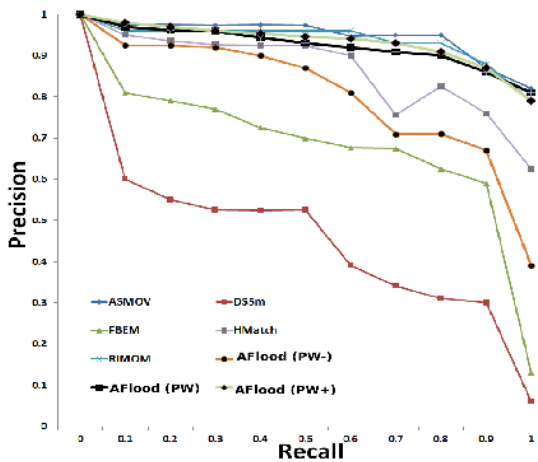

*Figure 4: Instance matching results against IIMB benchmarks. In the figure, instance matcher AFlood (PW-) denotes our core instance matcher without property weight, AFlood (PW) denotes our previous instance matcher considering primitive weight factors and AFlood (PW+) is the instance matcher with proposed property weight metric.*

## 6.5. Result of Our Proposed System with IIMB-2009 Data Set

Table 3 shows the results of different transformations when our proposed metric of automatic weight generation is considered. The results depict that the proposed metric of automatic weight generation for properties has a positive impact in instance matching technique.

Table 3. Instance matching results of different transformations when our proposed metric of automatic weight generation.

| Datasets | Transformation | Prec. | Rec. | F-Measure |
|---|---|---|---|---|
| 001-010 | Value transformations | 1.00 | 1.00 | 1.000 |
| 011-019 | Structural transformations | 0.91 | 0.84 | 0.868 |
| 020-029 | Logical transformations | 1.00 | 1.00 | 1.000 |
| 030-037 | Several combinations of the previous transformations | 0.96 | 0.83 | 0.885 |

As a summation on IIMB 2009 data sets, Fig. 4 shows the recall-precision graph depicting our three different approach of instance matching system: 1. our core instance matcher without property weight, called as *AFlood (PW-);* 2. our augmented instance matcher with primitive property weight, we called as *AFlood (PW);* and 3. our further improved instance matcher with proposed automatic weight factor, we are calling as *AFlood (PW+).* The figure depicts the improvement of our three different instance matcher.

## 6.6. Result of Our Proposed System with IIMB-2010 Data Set

In the instance matching track of OAEI-2010[4], the participants for IIMB2010 large dataset were Combinatorial Optimization for Data Integration (CODI) [20], Automated Semantic Mapping of Ontologies with Validation (ASMOV) [35] and RiMOM [19]. We experimented with our core instance matching system without property weight, we called as *AFlood(PW-)*, and our augmented instance matching system with proposed automatic weight factor, we call as *AFlood(PW+)*. Fig.5 shows the recall-precision graph of the participants [34] and the curves of our instance matching systems. Our proposed method outperforms other methods in several cases although CODI shows better result in some cases.

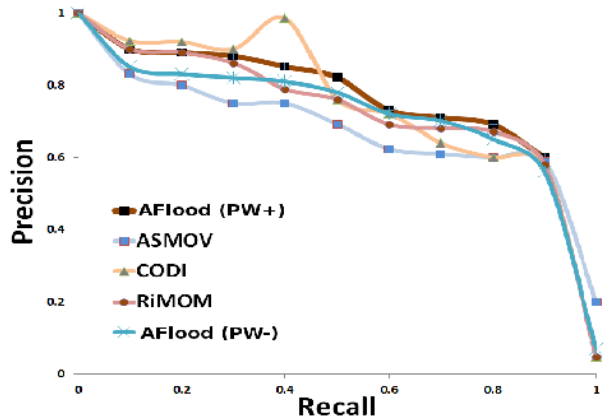

*Figure 5: Instance matching results against IIMB 2010 datasets. In the figure, our instance matcher, called as AFlood (PW+), shows the effectiveness of considering our proposed automatic weight generation factors*

---

4      http://oaei.ontologymatching.org/2010/

## 7. CONCLUSIONS

In this paper, we address a unique idea of generating non-linear property weights automatically in instance matching technique to integrate semantically rich data, often called as instances of semantic knowledge base. Our mathematical reasoning section logically satisfies the theoretical strength of the proposed method from different aspects. We mathematically model a metric for generating property weights in a knowledge base. The metric is then used in our instance matching algorithm to produce better results. Experiment and evaluation section exhibits how theoretically proven approach strongly contributes in achieving better outcome to integrate semantic data within same or among heterogeneous data sources. Our instance matcher with property weight provides better outcome than without property weight. Therefore, we can clearly state that automatic property weight generation in instance matching algorithm plays a vital role in semantic data integration. Application of this method in other domain such as record linkage, entity resolution problem, identity recognition may also open a new research scope.

Our future task covers to improve the scalability issues of the proposed method. Moreover, we would like to apply this integrator in integration of different social network data for investigating its applicability in real world.

**Authors**

**Md. Hanif Seddiqui** received his B.Sc.Eng. degree in Electronic and Computer Science from Shahjalal University of Science and Technology, Sylhet, Bangladesh in 2000 and his M.Eng. and D.Eng. degree in Computer Science from Toyohashi University of Technology, Japan in March 2007 and in March 2010 respectively. He is currently working as an Associate Professor at the Department of Computer Science and Engineering, University of Chittagong, Bangladesh. His current research interest includes Ontology Alignment, Knowledge Engineering, Bioinformatics and Semantic Web Techniques in Information Retrieval and Big Data. 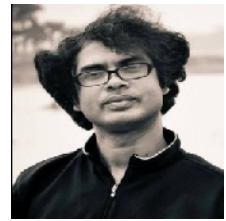

**Rudra Pratap Deb Nath** received the B.Sc. degree in Computer Science and Engineering in 2010 and M.Eng. degree in Computer Science and Engineering from Toyohashi University of Technology, Japan in 2013. He is an Assistant Professor at the Department of Computer Science and Engineering, University of Chittagong, Bangladesh. His research interests include Data integration, Knowledge Engineering, and Semantic Web Techniques in Information Retrieval and Data Mining. 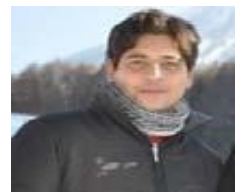

**Masaki Aono** received his B.Sc. and M.Sc. degree in Information Science from Faculty of Science, University of Tokyo in March, 1981 and in March 1984 respectively and his Ph.D. degree in Computer Science from Rensselaer Polytechnic Institute in May, 1994. He is currently working as a Professor at the Department of Computer Science and Engineering, Toyohashi University of Technology, Japan. Masaki Aono is a member of ACM (Association for Computing Machinery), IEEE Computer Society, IPSJ (Information Processing Society of Japan), IEICE (The Institute of Electronics, Information and Communication Engineers), JSAI (The Japanese Society for Artificial Intelligence), and NLP (Natural Language Processing Society). His current research interest includes rtificial Intelligence, Signal, Image and Video Processing, Data Mining and Machine Learning 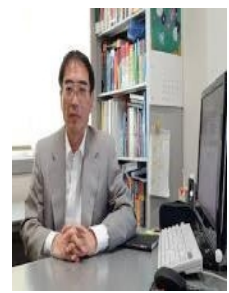